\documentclass[10pt,twocolumn,letterpaper]{article}

\usepackage{btas}
\usepackage{times}
\usepackage{epsfig}
\usepackage{graphicx}
\usepackage{subfigure}
\usepackage{epstopdf}
\usepackage{amsmath}
\usepackage{amssymb}
\usepackage{multirow}
\usepackage{url}
\usepackage{booktabs}

\btasfinalcopy 

\ifbtasfinal\fi

\begin{document}

\title{Feature Map Pooling for Cross-View Gait Recognition \\
	Based on Silhouette Sequence Images}

\author{Qiang Chen, Yunhong Wang, Zheng Liu, Qingjie Liu\thanks{corresponding author}
\ and Di Huang\\
Beijing Advanced Innovation Center for Big Data and Brain Computing, Beihang University\\
Beijing, China\\
{\tt\small \{chenqiangcq, yhwang, zhengliu, qingjie.liu, dhuang\}@buaa.edu.cn}
}

\maketitle
\thispagestyle{empty}
\begin{abstract}
In this paper, we develop a novel convolutional neural network based approach to extract and aggregate useful information from gait silhouette sequence images instead of simply representing the gait process by averaging silhouette images. The network takes a pair of arbitrary length sequence images as inputs and extracts features for each silhouette independently. Then a feature map pooling strategy is adopted to aggregate sequence features. Subsequently, a network which is similar to Siamese network is designed to perform recognition. The proposed network is simple and easy to implement and can be trained in an end-to-end manner. Cross-view gait recognition experiments are conducted on OU-ISIR large population dataset. The results demonstrate that our network can extract and aggregate features from silhouette sequence effectively. It also achieves significant equal error rates and comparable identification rates when compared with the state of the art.
\end{abstract}

\section{Introduction}

Biometrics such as the face, iris, fingerprint and signature are widely applied for human identity authentication. One major limitation of these biometrics is that we need subject cooperation to acquire these biometrics, which is difficult to be implemented in an uncooperative environment. Gait, as an important biometric cue, overcome this limitation since it can be easily captured by a surveillance camera at long distance in uncontrolled scenarios without subject cooperation.

Nowadays, a large number of surveillance cameras are installed in almost every corner of cities, such as shopping malls, hotel, airports, rail stations, etc. Cameras provide a large volume of useful data for crimes and forensic identification. Among the techniques used in surveillance technology and forensic identification, gait recognition is one of the most powerful methods. It has already been applied in a real case to convict criminals~\cite{bouchrika2011using}. However, gait recognition is still
a challenging task due to large variations in walking speeds, clothing, viewpoints and carrying conditions. A lot of methods were proposed
to solve these problems. Most of these methods can be grouped into model-based and appearance-based approaches~\cite{wu2016comprehensive}. 

The model-based approaches try to build models to reconstruct underlying structures of the human body from video sequences. For example, in ~\cite{bobick2001gait} and~\cite{johnson2001multi}, they used four parameters, including the height of the body, the distance between head and the pelvis, the distance between the pelvis and left foot, the distance between pelvis and right foot and the distance between left and right foot, to represent the structure of a body. Then gait recognition is completed based on these four parameters.
Ariyanto et al.~\cite{ariyanto2011model} used 3D gait data reconstructed from multiple cameras~\cite{seely2008university} to perform recognition. View variation may not be an issue when multiple cameras are available. 3D data conveys more information than 2D data, thus can achieve high accuracy. However, 3D data acquisition costs a lot and should be conducted in a controlled environment, which limits its application. 

The appearance-based approaches take surveillance image sequences as input
instead of modeling the underlying structure of the human body. To reduce the impact of clothing, silhouette based representation is prevalent within the gait recognition community~\cite{lam2011gait, liu2004simplest, man2006individual}. The first step of silhouette based representation is extracting a binary silhouette sequence from a video sequence. Then several methods can be used to aggregate gait silhouette sequence into one image. Gait Energy Image 
(GEI)~\cite{man2006individual} is one of the most popular representation and it is obtained by averaging silhouette sequence over a complete gait cycle(s). Although only one single image is generated, GEI encodes spatial and temporal information of a gait cycle, thus achieves promising results. Based on GEI, various approaches have been proposed to enhance the performance of gait recognition. Tao et al.~\cite{tao2007general} proposed Gabor features which are obtained by convolving the averaged
gait image with Gabor filters. Xu et al.~\cite{xu2012human} proposed a patch distribution feature  which representes each GEI as a set of local augmented Gabor features. Similarly, Guan and Li~\cite{guan2013robust} convolved GEI with Gabor filters from five scales and eight orientations to generate Gabor-GEI feature template. 

Other features are also developed to represent motion or/and appearance information of gait silhouette sequences. Inspired by the Motion History Image (MHI)~\cite{bobick2001recognition} which was developed for human action recognition, Lam and Lee~\cite{lam2005new} proposed Motion Silhouettes Image (MSI) to embed spatial and temporal information of gait silhouettes. Later in \cite{lam2011gait}, Lam introduced Gait Flow Image (GFI) for gait recognition. Bashir et al.~\cite{bashir2009gait}
proposed Gait Entropy Image (GEnI), which captures most motion information and encodes the information in a single image.

The most intractable problem in gait biometric is cross view gait recognition which has been being a hot research direction for years. 
Numerous studies have made great efforts to tackle this problem. As mentioned above, model-based methods especially 3D model based methods are good solutions to this despite of high cost. Appearance-based methods either focus on extracting view-invariant gait features or project extracted features from different viewpoint to a subspace which minimizing the variance of view-change~\cite{chen2014cross,shiraga2016geinet, wu2016comprehensive,zheng2011robust}. Reviews on gait recognition can be found in \cite{connie2015review} and \cite{prakash2016recent}.

Deep learning has been successfully applied in many computer vision tasks, such as image classification~\cite{krizhevsky2012imagenet}, video classification~\cite{karpathy2014large}, human pose estimation~\cite{bulat2016human} and face recognition~\cite{taigman2014deepface}. In these research areas, deep learning methods, especially Convolutional Neural Networks (CNN), accomplish significant progress by learning rich features from large volumes of training data.

Many CNN-based methods~\cite{shiraga2016geinet,wu2016comprehensive,zhang2016siamese} were explored to perform gait recognition, which also achieves remarkable improvements. CNN can automatically extract hierarchal features from given image, which is far more efficient than hand-crafted features. In addition to feature extraction, deep learning based similarity measuring methods have also been proposed. Among which, Siamese network is the most popular one. The Siamese network architecture is a
useful tool to learn similarity metric between a pair of inputs by learning sufficient feature representations that make inter-class distance close while intra-class distance large~\cite{chopra2005learning,taigman2014deepface}. 

A Siamese neural network~\cite{chopra2005learning} contains two parallel branches sharing the same weights. In training stage, pairs of similar and dissimilar data are fed into the two branches separately. Then the outputs from two branches are combined by matching layers to compute the contrastive loss. Back propagation algorithm is used to train the model. In the testing stage, the Siamese network calculates the distance between the query input
and every gallery data, and choose the closest gallery as result. In \cite{wu2016comprehensive} and  \cite{zhang2016siamese}, Siamese networks are applied on gait recognition. The inputs of these neural networks are GEI and CGI~\cite{wang2010chrono} respectively. The Siamese neural network achieves two purposes: the first is extracting features from the input image; the second is mapping features to the target space defined by the specific task. 
In this work, we argue that we can learn useful
information from raw data, \ie binary silhouette sequence, directly, and fuse them in feature level instead of data level. To achieve this, we propose an improved Siamese neural network that learns features directly from raw silhouette sequence images and fuses them in a layer-wise pooling way. Subsequently, additional convolutional layers are applied to map the fused features into task space. The method can be used to cross-view gait recognition. We test it on OU-ISIR large population
dataset~\cite{iwama2012isir}, and obtain promising performance which is comparable with the state of art. 


\section{Proposed method}
\begin{figure}[!t]
	\begin{center}
		\includegraphics[width=0.8\linewidth]{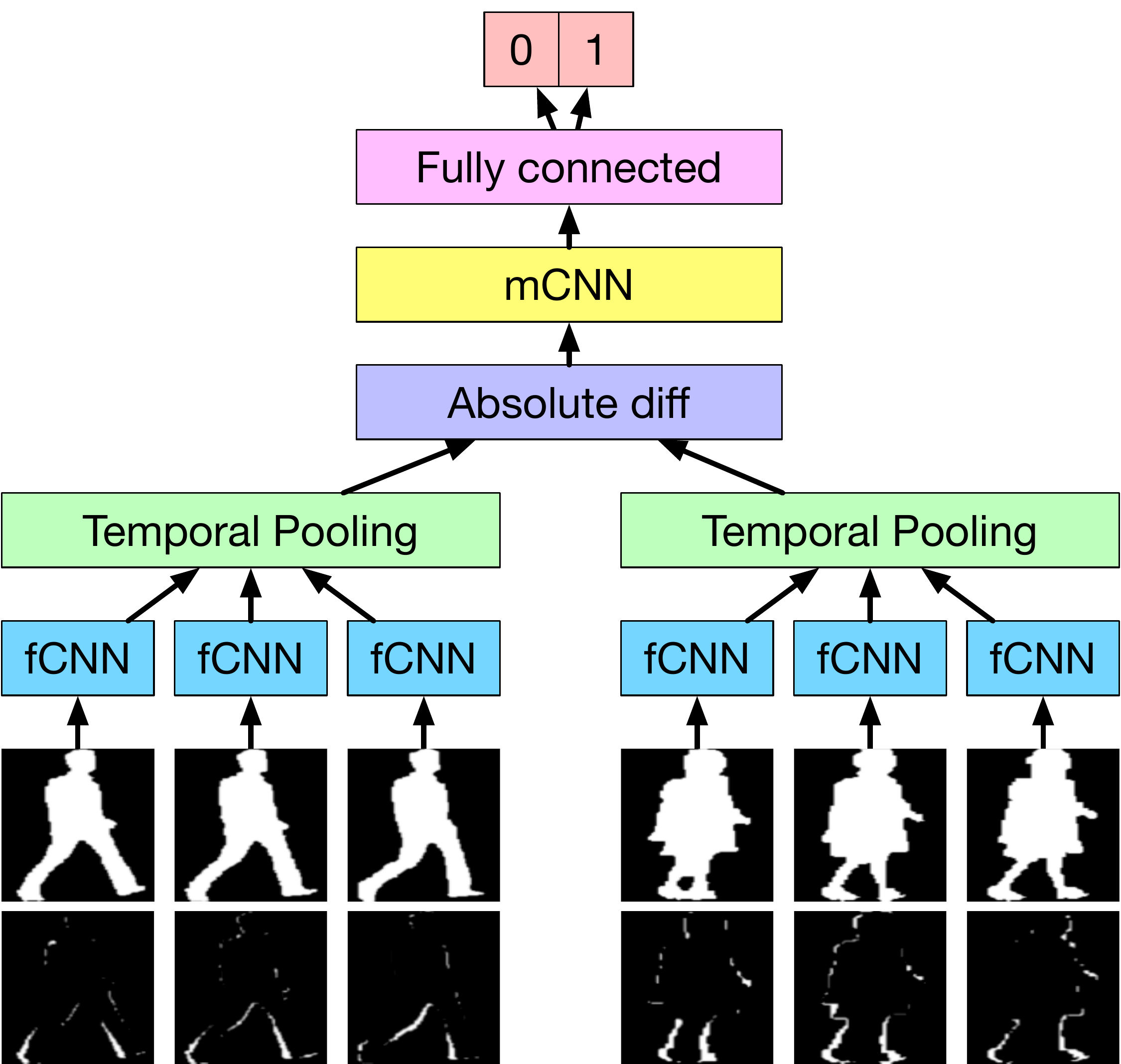}
	\end{center}
	\caption{The network architecture of our method. Each silhouette image and the difference image between two adjacent silhouettes are fed into the fCNN to extract features. Feature map pooling is used to aggregate outputed feature maps from fCNNs into a single fixed size one. Absolute difference module calculates element-wise differences between two aggregated feature maps. mCNN is the second convolutional neural network that project absolute difference feature maps into task space. Fully
    connected module outputs the similarity distribution of given two sequences.}
	\label{fig:network}
\end{figure}
\begin{figure*}[!t]
	\begin{center}
		\includegraphics[width=0.8\linewidth]{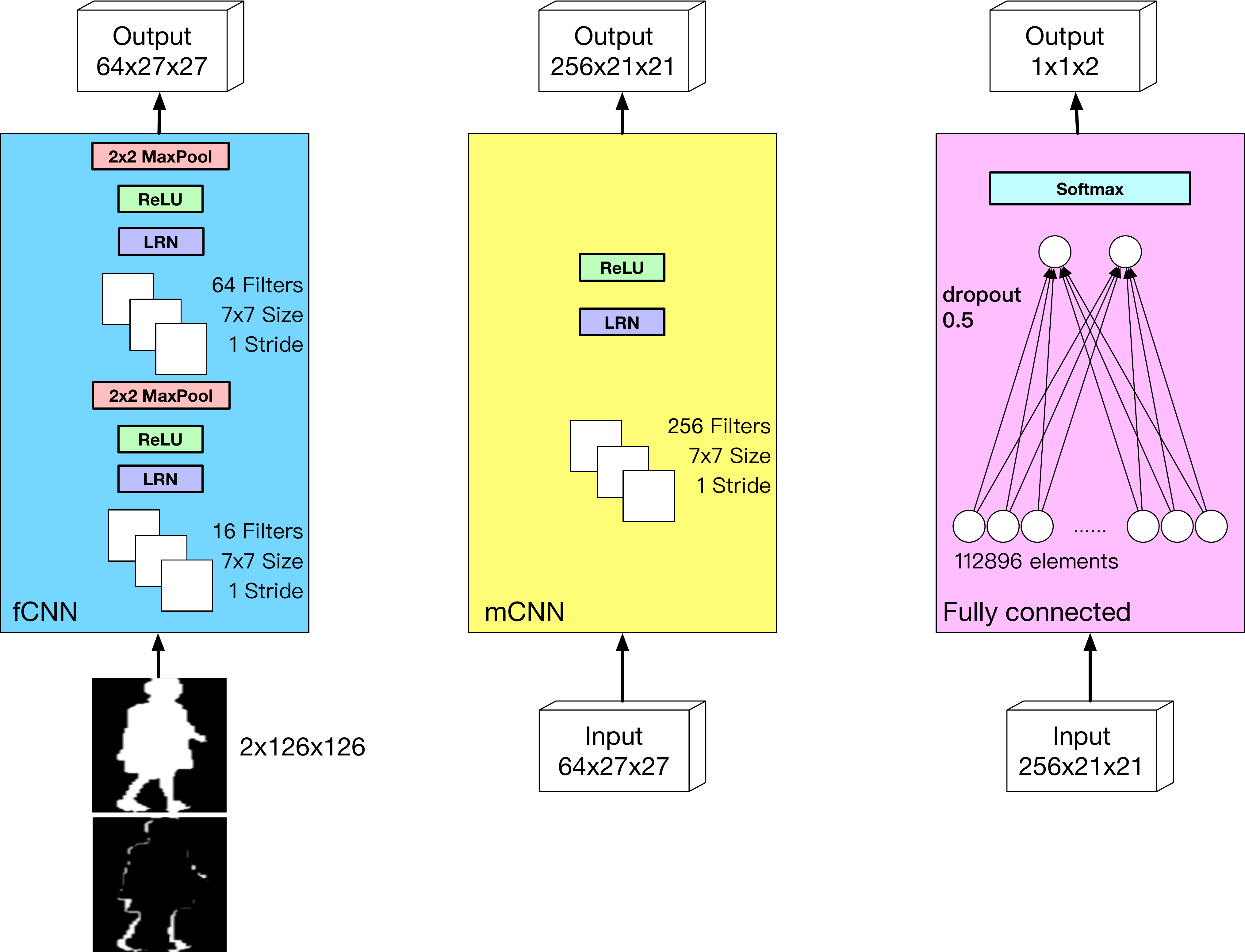}
	\end{center}
	\caption{Details of fCNN, mCNN and fully connected layers.}
	\label{fig:network_detail}
\end{figure*}
The brief architecture of our method is shown in Figure~\ref{fig:network}. Each silhouette and the difference of two adjacent frames are fed to a convolutional neural network to extract useful features representing gait information at the moment (we call it fCNN for short). Inspired by the spatial pooling within a feature layer and the fact that adjacent frames are highly correlated, we explore using a layer-wise pooling method to fuse outputs of fCNN. Layer-wise pooling can
convert arbitrary length gait sequences into fixed size feature maps which preserving spatial and temporal gait information. The followed is a classical Siamese network architecture. Given a pair of sequences, corresponding  feature maps are obtained through fCNN and layer-wise pooling. Then, the absolute difference is computed between the two fixed feature maps. A one-layer CNN is used to map the difference features into a vector (we call it mCNN for short). It should be noted that more layers
are feasible. A fully connected module will convert this vector into two probabilities indicating the pair inputs are the same or not.

To predict the identity of a given probe sample, the similarities between probe gait sequences and every gallery gait sequences are inferred by the whole network. Then the identity of the most similar gait gallery is chosen as the probe sample's identity.
\subsection{fCNN}

As shown in Figure \ref{fig:network}, the silhouette image and the difference image between two silhouettes are processed by fCNN. The detailed parameters of fCNN are shown in the left part of Figure~\ref{fig:network_detail}. fCNN contains two convolutional layers. The first one includes 16 filters. The second one includes 64 filters. All filters are of size 7$\times$7 and applied with one stride. Spatial pooling and local response normalization (LRN) are appended after each convolutional layer.
The spatial pooling operations are applied in 2$\times$2 neighborhood. The LRN arguments are set to values recommended from\cite{krizhevsky2012imagenet}. After applying the first convolutional layer and pooling layer, we obtain 16 feature maps sized $60 \times 60$, and 64 $ 27 \times 27$ feature maps after the second convolutional and pooling layer. For notational simplicity, we refer to fCNN as a function $f =\mathrm{fCNN}(x)$, which takes a silhouette gray image
and difference image $x$ as input and produces feature maps $f$ as output. The size of $f$ is 64$\times$27$\times$27.  Let $s = s^{(1)} ...s^{(T)}$ be the input sequence data of length T, where one channel of $s^{(t)}$ is the image at time t and another channel is the difference between image at t and image at t-1. The silhouette
 image at time 0 is ignored because there is no previous image. It should be noted the layer-wise pooling introduced in the following subsection can fuse arbitrary frames, so the length T is not fixed. Each silhouette $s^{(t)}$ will go through the fCNN to produce feature maps, $f^{(t)} = \mathrm{fCNN}(s^{(t)})$.

\subsection{Feature map pooling}
One straightforward way to tackle temporal sequence is using recurrent neural networks to encode information across time~\cite{mclaughlin2016recurrent}. Another widely used method is 3D CNN, which is developed by \cite{ji20133d} to perform action recognition. Inspired by the spatial pooling used in CNN and the fact that adjacent frames in a video are highly redundant, we proposed to use feature map pooling to aggregate extracted features from sequence frames. 

Feature maps for each frame can be obtained through fCNN. Similar to spatial pooling, there are two ways to aggregate these features, max pooling and mean pooling. 
\begin{equation}
    v{(k, h, w)}=\max \limits_{t = 1,...,T} f^{(t)}(k, h, w)
\end{equation}
where $v(k, h, w)$ is the value of $k$th fused feature map at position $(h, w)$, and $f^{(t)}(k, h, w)$
is value of $k$th feature map of frame $t$ at $(h, w)$. Finally, arbitrary number feature maps are merged into one whose size is 64$\times$27$\times$27.

mean pooling is also a commonly used aggregation strategy. It is used here to produce a single feature maps averaged over all the extracted feature maps, 
as follows:
\begin{equation}
v{(k, h, w)}=\frac{1}{T}\sum \limits_{t = 1,...,T} f^{(t)}(k, h, w)
\end{equation}

In this paper, we test both pooling methods. 
But we found that mean pooling performance is around 5\% worse than max pooling
when comparing cross-view identification precision average in experiments, so max pooling was chosen in our experiments. 

\subsection{Similarity measurement}
Given a pair of fused feature maps for two sequences, the task is to identify whether the two sequences represent the same person. To this end, a network similar to Siamese neural network is employed to measure the similarity between two fused feature maps. Firstly, the absolute difference between two fused feature maps is obtained. The output difference feature maps have size same to the fused map, which is 64$\times$27$\times$27. Then a mapping convolutional layer, \ie mCNN, is
applied to project the difference feature to a similarity vector. mCNN is a one layer convolutional network which has 256 filters sized 7$\times$7. The detail of mCNN module is shown at the middle part of Figure \ref{fig:network_detail}. The output size of mCNN is 
256$\times$21 $\times$21. Then it is reshaped to one dimension vector with 112896 elements. The following fully connected layer take this vector as input and produce the final result. Detail information of fully connected module is shown at the right part of Figure \ref{fig:network_detail}.  

\section{Experiments}

\subsection{Dataset}

\begin{figure*}[!ht]
	\begin{center}
		\includegraphics[width=1\linewidth]{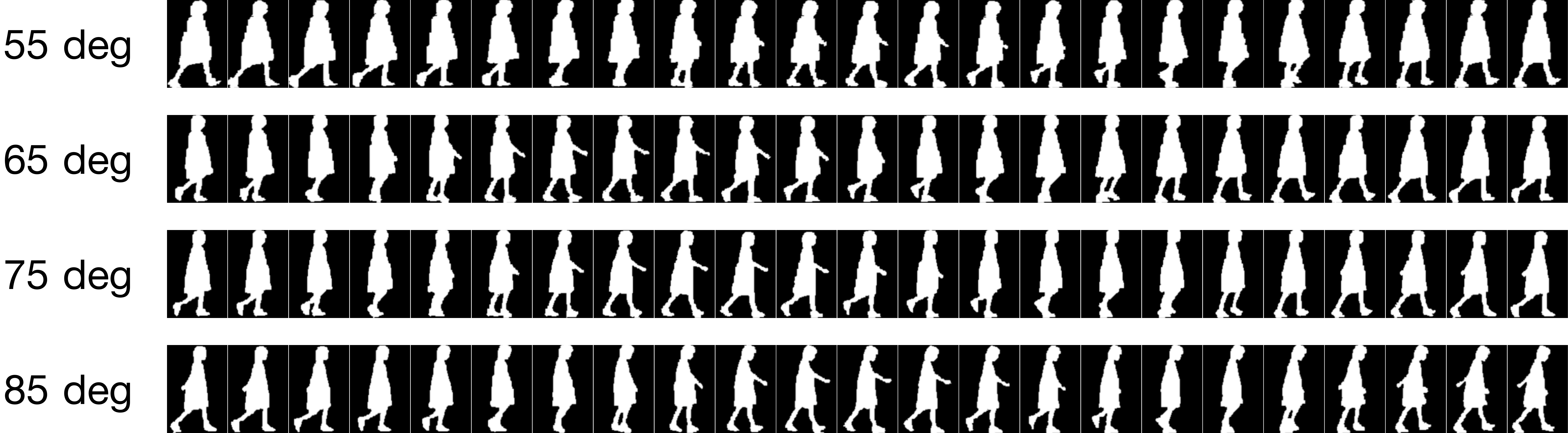}
	\end{center}
	\caption{
		Probe silhouette sequences of OULP-C1V1-6218964. 
		Each identity has two sequence subsets: probe subset and gallery subset. And each subset has four sequences with different view angles: 55 degree, 65 degree, 75 degree and 85 degree.}
	\label{fig:OULP_example}
\end{figure*}

We test our method on OU-ISIR large population dataset~\cite{iwama2012isir}, as it is the largest gait dataset suitable for training deep neural networks. There are two versions for OU-ISIR large population dataset: OULP-C1V1 and OULP-C1V2~\footnote{http://www.am.sanken.osaka-u.ac.jp/BiometricDB/GaitLP.html}. OULP-C1V1 contains 4,007 subjects, while OULP-C1V2 includes 4,016 subjects. Aside from this difference, OULP-C1V2 has a more accurate bounding box for each silhouette and the size of
moving-average filter applied in the size-normalized silhouette creation process. In this work, the first version of OU-ISIR, OULP-C1V1, was used to evaluated the performance of our method. Figure ~\ref{fig:OULP_example} shows the full silhouette images of subject OULP-C1V1-6218964 with four observation views: 55, 65, 75 and 85 deg.

In this work, we follow protocol used in \cite{mansur2014cross}, only a subset of OULP-C1V1 is used to test our method. The subset contains 1912 subjects. And each subject has probe and gallery gait sequences with different angle views: 55, 65, 75 and 85 degree. There are 8 sequences for one subject. The length of sequence ranges from 19 to 43 frames.

\subsection{Training}

The network inputs are pairs of arbitrary long gait sequences. Each silhouette is resized to 126$\times$126. In each mini batch, half of the input pairs have same identities. For one probe sequence, we pick its corresponding gallery sequence with a random view angle to form a positive training sample pair. Similarly, another gallery sequence with different identity can be selected to form negative training sample pair. 

We use negative $\ln$ loss and stochastic gradient descent to train our network. 
\begin{equation}
    loss=-(t_0 \ln p_0 + t_1 \ln p_1)
\end{equation}

\begin{figure}[!ht]
	\begin{center}
		\includegraphics[width=1\linewidth]{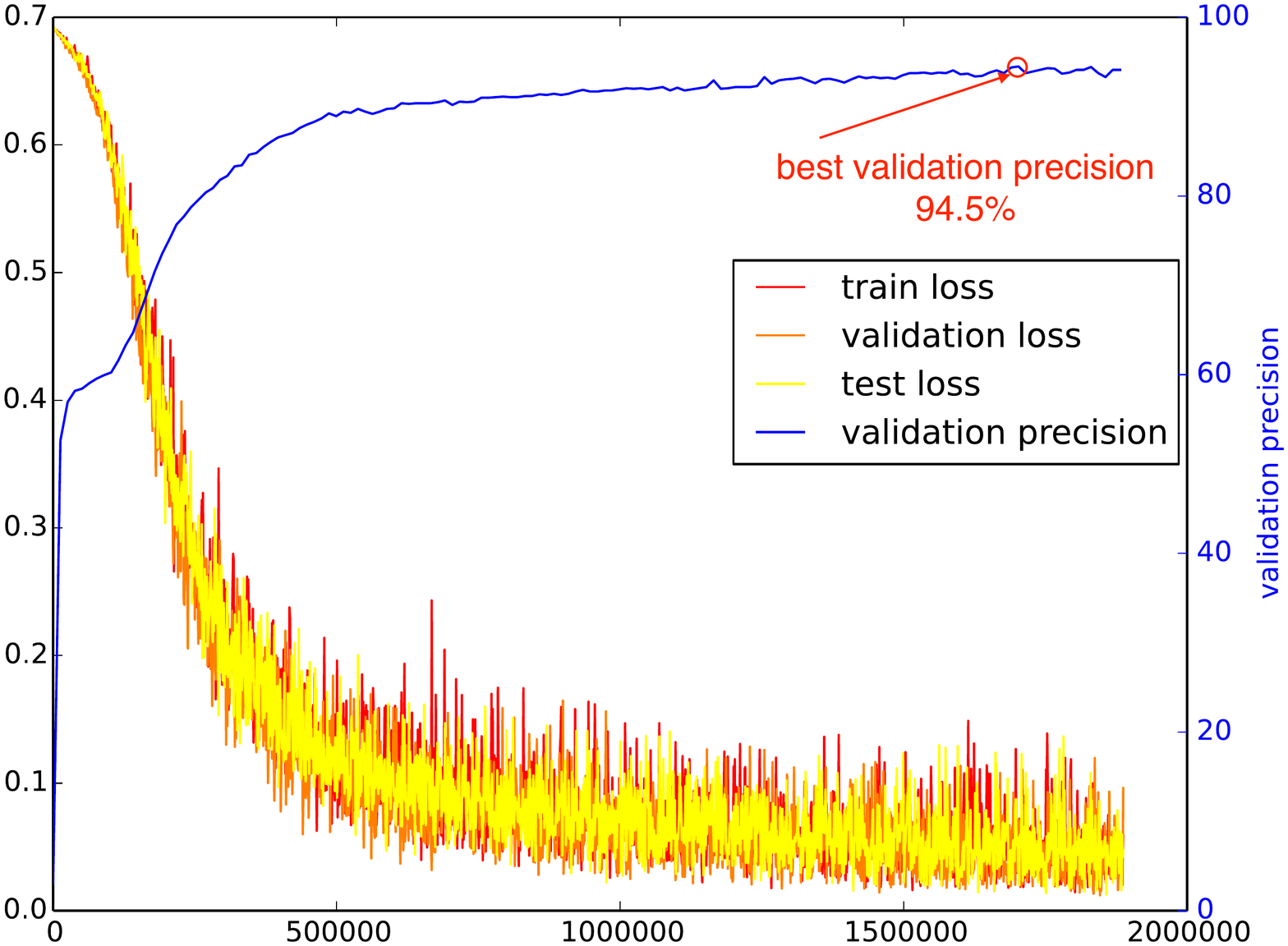}
	\end{center}
	\caption{The network was trained with about 1.8 million iterations. The validation precision achieves $94.5\%$. }
	\label{fig:losscurves}
\end{figure}

1912 subjects are divided into two groups with the same size for training and testing without overlapping\footnote{http://www.am.sanken.osaka-u.ac.jp/\~{}mansur/files/list\_train\_test.txt}. Images are resized to 126$\times$126 to input the networks. We randomly select 100 subjects in training set as validation set, so 856 subjects are left for training. It should be noted that no data augmentation is used during training. The size of mini-batches was set to 128, learning rate was set to
0.001, momentum was set to 0.0. LRN
was set to default as suggested in \cite{krizhevsky2012imagenet}. The networks were written in Torch 7 and trained on a NVIDIA GeForce GTX Titan X. We run validation test every 100 iterations. It will cost several minutes. The number of iterations is up to 1.8 million and the training phase lasted 7 days. The model with highest recognition rate was chosen for evaluation. Figure~\ref{fig:losscurves} shows the loss curves and precision increasing with respect to the number of iterations.

Given a probe gait sequence and a gallery gait set, the similarity between the 
probe sequence and each sequence in the gallery set are evaluated by trained network. The identity of the probe gait sequence will be assigned to the most similar one in the gallery. For cross-view recognition, 16 recognition tasks need to be done on different probe view and gallery view setting, $\{55, 65, 75, 85\} \times \{55, 65, 75, 85\}$. It is very slow to calculate cross view gait recognition rate on test set with
956 subjects. There are $956 \times 956 = 913936$ measurements should be computed between subjects for each task. To facilitate calculation, we storage the fused features for each silhouette sequence and similarities only be calculated between fused features. This will drastically reduce computing time. 16 cross view recognition tasks can be done within 5 hours on GPU.
\subsection{Impact of sequence length}
\begin{figure*}[!htb]
	\centering
	\subfigure[]{
		\label{fig:subfig:pre_c} 
		\includegraphics[width=0.4\linewidth]{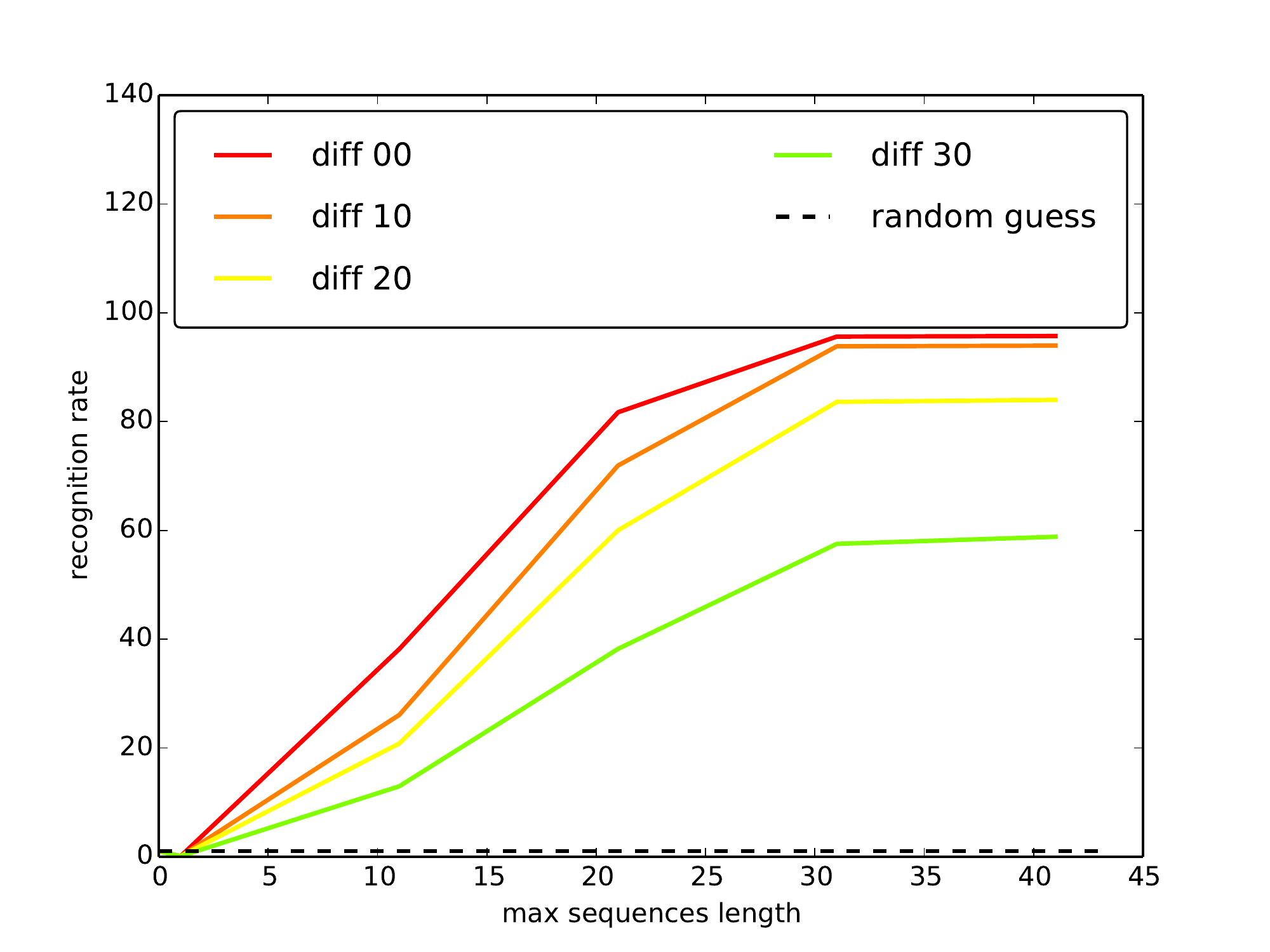}}
	\subfigure[]{
		\label{fig:subfig:eer_c} 
		\includegraphics[width=0.4\linewidth]{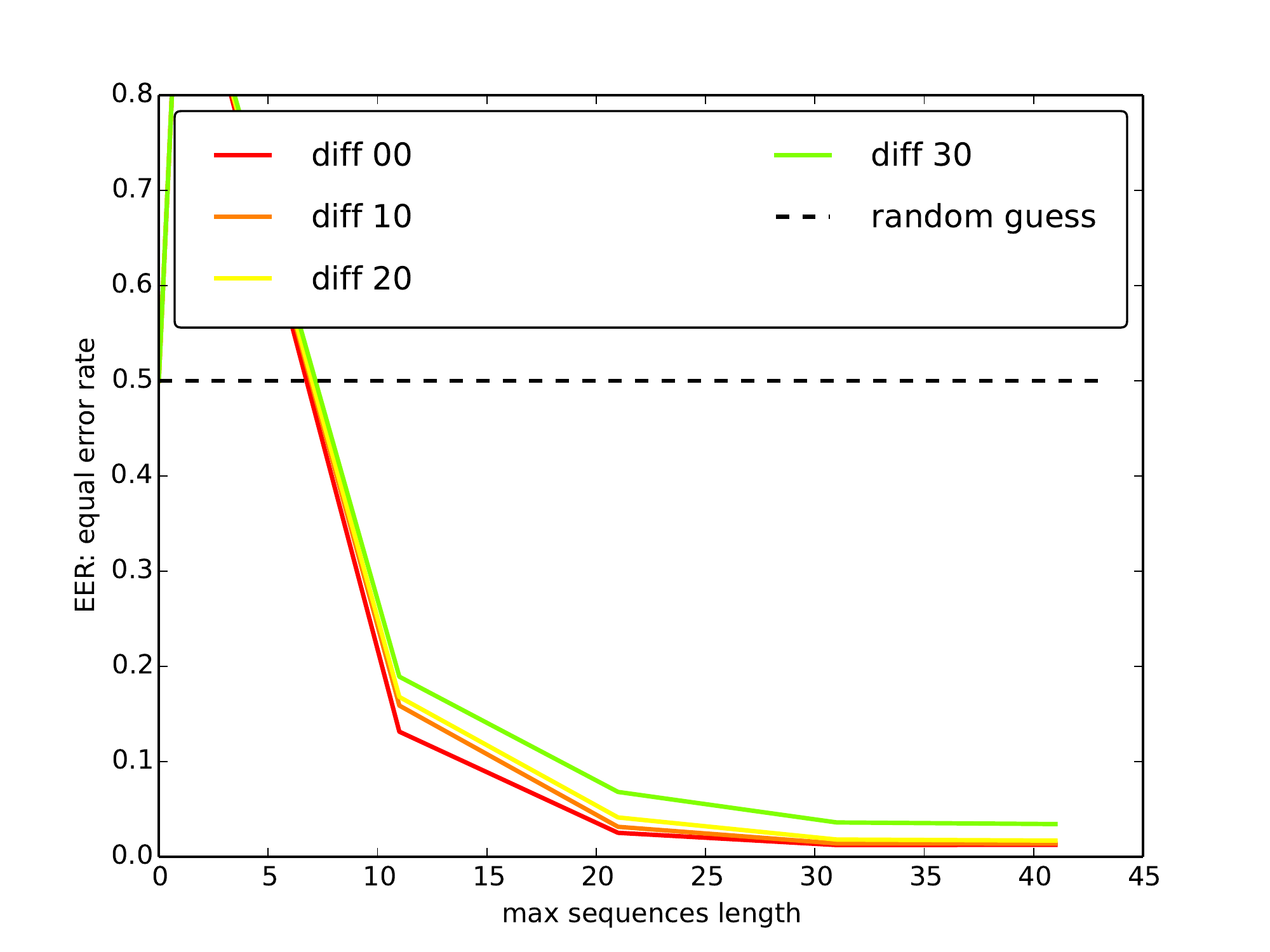}}
	\caption{Influence of sequence length on recognition rates and EERs.}
	\label{fig:pre_candeer_c} 
\end{figure*}

Our method can take arbitrary length sequence as input. However, a longer sequence may improve the performance of recognition. To evaluate this, we conduct experiments on different sequence length ranging from 1 frame to 43 frames. The results are shown in Figure~\ref{fig:subfig:pre_c} and  Figure~\ref{fig:subfig:eer_c}. The precision and EERs are averaged across different view angles between probe and gallery sets.

\subsection{Results}
\begin{table*}[!h]
	\begin{center}
		\begin{tabular}{ccccccccccccccccc}
			\toprule
			\multirow{3}{*}{ Probe } &  \multicolumn{15}{c}{Gallery} \\
			\cmidrule(lr){2-16} 
			&\multicolumn{5}{c}{Rank-1} & \multicolumn{5}{c}{Rank-2} &\multicolumn{5}{c}{Rank-5} \\
			\cmidrule(lr){2-6} \cmidrule(lr){7-11} \cmidrule(lr){12-16}
			& 55 & 65 & 75 & 85 & Mean & 55 & 65 & 75 & 85 & Mean & 55 & 65 & 75 & 85 & Mean \\
			\midrule   
			55 & 95.2 & 93.6 & 81.2 & 62.2 & 83.1 & 97.1 & 96.3 & 89.3 & 74.5 & 89.3 & 98.2 & 98.2 & 95.5 & 87.8 & 94.9 \\
			65 & 90.9 & 95.3 & 95.5 & 90.2 & 93.0 & 94.8 & 97.7 & 97.7 & 94.6 & 96.2 & 97.6 & 98.4 & 98.8 & 97.6 & 98.1 \\
			75 & 77.5 & 94.4 & 96.0 & 94.2 & 90.5 & 87.2 & 96.9 & 98.3 & 97.4 & 95.0 & 93.6 & 98.7 & 99.0 & 98.5 & 97.5 \\
			85 & 55.4 & 87.1 & 94.8 & 94.7 & 83.0 & 68.9 & 93.7 & 97.0 & 97.9 & 89.4 & 83.3 & 96.8 & 98.5 & 98.6& 94.3 \\
			\bottomrule
		\end{tabular}
	\end{center}
	\caption{Performance of our method on OULP-C1V1 following protocol used in \cite{mansur2014cross} in terms of Rank-1, Rank-2 and Rank-5 recognition rates.}
	\label{tab:rankrec}
\end{table*}

We follow \cite{mansur2014cross} protocol to test our method. Only a subset of 856 subjects is used for training. Evaluation is conducted on 956 subjects. Table~\ref{tab:rankrec} reports the performance of our method in terms of Rank-1, Rank-2 and Rank-5 recognition rates. Table~\ref{tab:eerandrecrate} lists equal error rates (EERs). From these two tables, we can see that the proposed method shows promising results on OULP-C1V1 gait dataset. To the best of our knowledge, there are no previous works reporting cross-view EERs and recognition rate fully, only EERs and recognition rate between 85 degree gallery and each 55, 65 and 75 degree were reported in this work \cite{mansur2014cross}. 

Furthermore, we compare our method with LDA~\cite{otsu1982optimal}, DATER~\cite{yan2005discriminant}, MvDA~\cite{mansur2014cross}, GMLDA~\cite{sharma2012generalized}, and CCA~\cite{liu2011joint} to demonstrate its superiority, as shown in Table \ref{tab:rec}. It can be seen that our method outperforms these methods significantly in terms of EERs which is an important verification indicator. Both GMLDA and MvDA require view information as input, thus achieve better identification rate than our
method. However, our method is blind to view angle information. Even though, it still performs well in terms of Rank-1 recognition rate. 

\begin{table*}[!ht]
	\begin{center}
		\begin{tabular}{cccccccccccc}
			\toprule
            \multirow{2}{*}{ Probe } & \multicolumn{5}{c}{EERs(\%)} \\
			\cmidrule(lr){2-6}
			& 55 & 65 & 75 & 85 & Mean \\
			\midrule			
            55 & 1.57 & 1.57 & 2.20 & 3.74 & 2.27 \\
            65 & 1.59 & 1.15 & 1.44 & 1.46 & 1.41 \\
            75 & 1.78 & 1.15 & 1.16 & 1.47 & 1.39 \\
            85 & 3.14 & 1.36 & 1.15 & 1.15 & 1.70 \\
			\bottomrule
		\end{tabular}
	\end{center}
	\caption{Cross-view EER}
	\label{tab:eerandrecrate}
\end{table*}

\begin{table*}[!htb]
	\begin{center}
		\begin{tabular}{ccccccccc}
			\toprule
            \multirow{2}{*}{ Method } & \multicolumn{4}{c}{EER(\%)} & \multicolumn{4}{c}{Rank-1} \\
			\cmidrule(lr){2-5} \cmidrule(lr){6-9} 
			& 55 & 65 & 75 & 85 & 55 & 65 & 75 & 85 \\
			\midrule			
                Ours                              & \textbf{3.74}& \textbf{1.46} & \textbf{1.47}& \textbf{1.15} & 62.23 & 90.16 & 94.24 &94.66 \\
				LDA\cite{otsu1982optimal}         & 8    & 5 			 & 4 & - & 56 & 91 & 96  & -\\
				DATER\cite{yan2005discriminant}   & 30   & 22 			 & 16& -& 10 & 29 & 65 & -\\
				GMLDA\cite{sharma2012generalized} & 12   & 9 			 & 5 & -& 68 & 82 & 95 & -\\
				MvDA\cite{mansur2014cross}        & 7    & 5 			 & 4 & -& \textbf{88} & \textbf{96} & \textbf{97} & -\\
				CCA\cite{liu2011joint}            & 21   & 13 			 & 8 & -& 52 & 81 & 92 & -\\
			\bottomrule
		\end{tabular}
	\end{center}
	\caption{Comparisons with other five methods in terms of EERs and Rank-1 identification rates.}
	\label{tab:rec}
\end{table*}

\section{Conclusion}
This paper present a novel CNN based gait recognition method. The proposed network architecture combines the advantage of convolutional neural network and Siamese network which evaluate similarity between two given arbitrary length silhouette sequences instead of GEI. Firstly, CNNs is used to extract features from each frame of sequence and the difference between previous frame. Inspired by the spatial pooling used within feature maps, a feature map pooling is employed to aggregate extracted features from different frames. Subsequently, a one layer CNN maps the difference of two fused features into task space. Finally, fully connected layers perform recognition.
Experiments for cross-view gait recognition on OU-ISIR large population dataset are conducted. Our method outperforms other methods significantly when compared with EERs. Specifically, it yielded approximately two times better than other methods in verification accuracy.

\section*{Acknowledgement}
This work was supported by The National Key Research and Development Plan (Grant No.2016YFC0801002)

\bibliography{submission_example}

\end{document}